\documentclass[conference]{IEEEtran}
\usepackage{blindtext, graphicx}
\usepackage[justification=centering]{caption}
\RequirePackage[singlelinecheck=off]{caption}
\usepackage{balance}
\ifCLASSINFOpdf
\else
\fi

\begin{document}
%
\title{Depression Scale Recognition from Audio, Visual and Text Analysis}

\author{\IEEEauthorblockN{Shubham Dham\IEEEauthorrefmark{1}, Anirudh Sharma\IEEEauthorrefmark{2} , Abhinav Dhall\IEEEauthorrefmark{3}}
\IEEEauthorblockA{Department of Computer Science and Engineering, Indian Institute of Technology (IIT) Ropar, India\\
Email: \IEEEauthorrefmark{1}2015csb1117@iitrpr.ac.in , \IEEEauthorrefmark{2}2015csb1007@iitrpr.ac.in , \IEEEauthorrefmark{3}abhinav@iitrpr.ac.in}
}


%


\maketitle

\begin{abstract}
Depression is a major mental health disorder that is rapidly affecting lives worldwide. Depression not only impacts emotional but also physical and psychological state of the person. Its symptoms include lack of interest in daily activities, feeling low, anxiety, frustration, loss of weight and even feeling of self-hatred. This report describes work done by us for Audio Visual Emotion Challenge (AVEC) 2017 during our second year BTech summer internship. With the increase in demand to detect depression automatically with the help of machine learning algorithms, we present our multimodal feature extraction and decision level fusion approach for the same. Features are extracted by processing on the provided Distress Analysis Interview Corpus-Wizard of Oz (DAIC-WOZ) database. Gaussian Mixture Model (GMM) clustering and Fisher vector approach were applied on the visual data; statistical descriptors on gaze, pose; low level audio features and head pose and text features were also extracted. Classification is done on fused as well as independent features using Support Vector Machine (SVM) and neural networks. The results obtained were able to cross the provided baseline on validation data set by 17\% on audio features and 24.5\% on video features.
\end{abstract}


%
\IEEEpeerreviewmaketitle

\section{Introduction}
Depression is a common mood disorder which affect the lives of the individual suffering from it. It is a world wide problem affecting innumerable lives. Depressed people are more prone to anxiety, sadness, loneliness, hopelessness, and are frequently worried and disinterested. They find it hard to concentrate on their work, communicate with people and even become introvert \cite{symp}. The may suffer from insomnia, restlessness, loss of appetite and may have suicidal thoughts. To be diagnosed with depression, the symptoms must be present for at least two weeks \cite{depressionarticle}. Hence detection of depression is a major issue. The present techniques rely on clinicians review of the patient in person. Those methods are subjective, done on interview and depend on reports by the patient. With rise in the depression, some automatic and reliable means of depression recognition is required. Efforts are being made in this direction to assess and detect depression through computer vision and machine learning. \\
Depression detection can be done through audio and video recordings of the patient. Depressed people behave different from normal people which  can be detected in audio and visual recordings of the patient. Studies have shown that depressed people tend to avoid eye contact, engage less in verbal communication, speak anxiously in short phrases and monotonously \cite{pedersen} \cite{fossi} \cite{waxer}.  \\
AVEC provides opportunity to teams from around the globe to come forth and develop a program for identifying depression using the dataset provided. Taking motivation from AVEC 2016 \cite{AVEC16}, our team participated in AVEC 2017 \cite{AVEC17}, which has similar dataset as AVEC 2016. Hence in this paper we present our  methods of visual, audio and text feature extraction, followed by decision level fusion which help identify depressed subjects with appreciable accuracy as evident in results obtained.

\section{Related Work}
Recent researches have been very prominent in detecting depressed people. Previous year's AVEC depression sub challenge have brought papers with decent accuracies as depicted in the results published. Le Yang et al. \cite{Le Yang} during AVEC 2016 achieved quite plausible accuracy through Decision Tree classification and multimodal fusion of audio-visual and text features. Asim Jan et al. \cite{asim} in AVEC 2014 competition used Motion History Histogram for depression recognition by extracting Local Binary Pattern (LBP) and Edge Orientation Histogram features (EOF). Low Lu-Shih  et al. \cite{acoustic} proposed description of depression recognition using acoustic speech analysis. The five acoustic feature categories used were prosodic, cepstral, spectral, glottal and Teager energy operator (TEO) based features, with TEO observed to perform best. The paper introduced benefits of Gaussian Mixture Model (GMM) clustering in depression recognition. Douglas Sturim  et al. \cite{sturim} also brought forward GMM clustering in classification on speech data.\\ 
Also other features such as head pose, blinking rate, and even textual features have been used for depression recognition. Head pose and facial movement is a discriminative feature to demarcate depressed people from non-depressed. Depressed people show less nodding \cite{fossi} , with more likeliness to position head down than non-depressed people \cite{waxer}, avoid eye contact \cite{fossi}. Alghowinem et al. \cite{sharifa} used head pose and movement features associated with the face to perform classification with SVM on depression recognition; and concluded that head movements of depressed people are different than that of normal person.\\
Inspired by the present deep learning performance in detecting emotion, Tzirakis et al. \cite{tzirakis} in AVEC 2016 presented a deep learning based approach in assessing emotion as well depression state of the person. They used Convolution Neural Network (CNN) on audio and Deep residual Network (resNet) of 50 layers on visual data. They concluded that their deep learning approach achieved relatively better results  than other methods at the time. Pampouchidou et al. \cite{pampou} in AVEC 2016 extracted features from the recorded verbal communication recorded in the transcript file provided in the data. Arousal-valence rating of the words which negative emotion were taken as a feature and  they concluded that its removal had a negative impact on the overall accuracy obtained on the trained fused model. 
Also Cummins et al. \cite{Cumm} during AVEC 2013, presented multimodal approach by fusing audio and visual modalities with different techniques to detect depression.\\
Hence all the recent researches are emphasizing on the need of depression recognition in some automatic and fruitful way to provide help to clinicians and patients. Therefore in this paper we also present our approach to detect the depression.

\section{Feature Extraction}

\subsection{Video Features}
 The challenge organizers did not share the raw video recordings, however 2-D, 3-D facial landmarks, Action Units (AUs), gaze and pose features were provided. Therefore we performed preprocessing on the 2-D facial landmarks and obtained features in the form of head  features, distance and blink rate. Statistics derived from the provided features of emotion, AUs, gaze, and pose were also used as complementary features.

\subsubsection{Head Features}
The Head pose features provided by the organizers were not used as they did not depict temporal information, i.e features did not convey the change with respect to time. Hence Head Motion was judged by horizontal and vertical motion of certain facial points, i.e., points {2, 4, 14, and 16} [Fig. 1] \cite{pampou}. The points selected were those that are minimally involved in facial movements and expressions such as blinking, smiling and other facial expressions, such that their motion prominently represent head motion.\\
For each of the above facial points, change in their position was calculated between every consecutive frame. This change was measured in both horizontal and vertical direction as well as net magnitude. Then statistical features were calculated as mean, median and mode of displacement in horizontal, vertical direction and magnitude of displacement, and velocity in horizontal, vertical direction and magnitude of velocity.

\captionsetup{singlelinecheck=on}
\begin{figure}[htp]
\centering
\includegraphics[width=8cm]{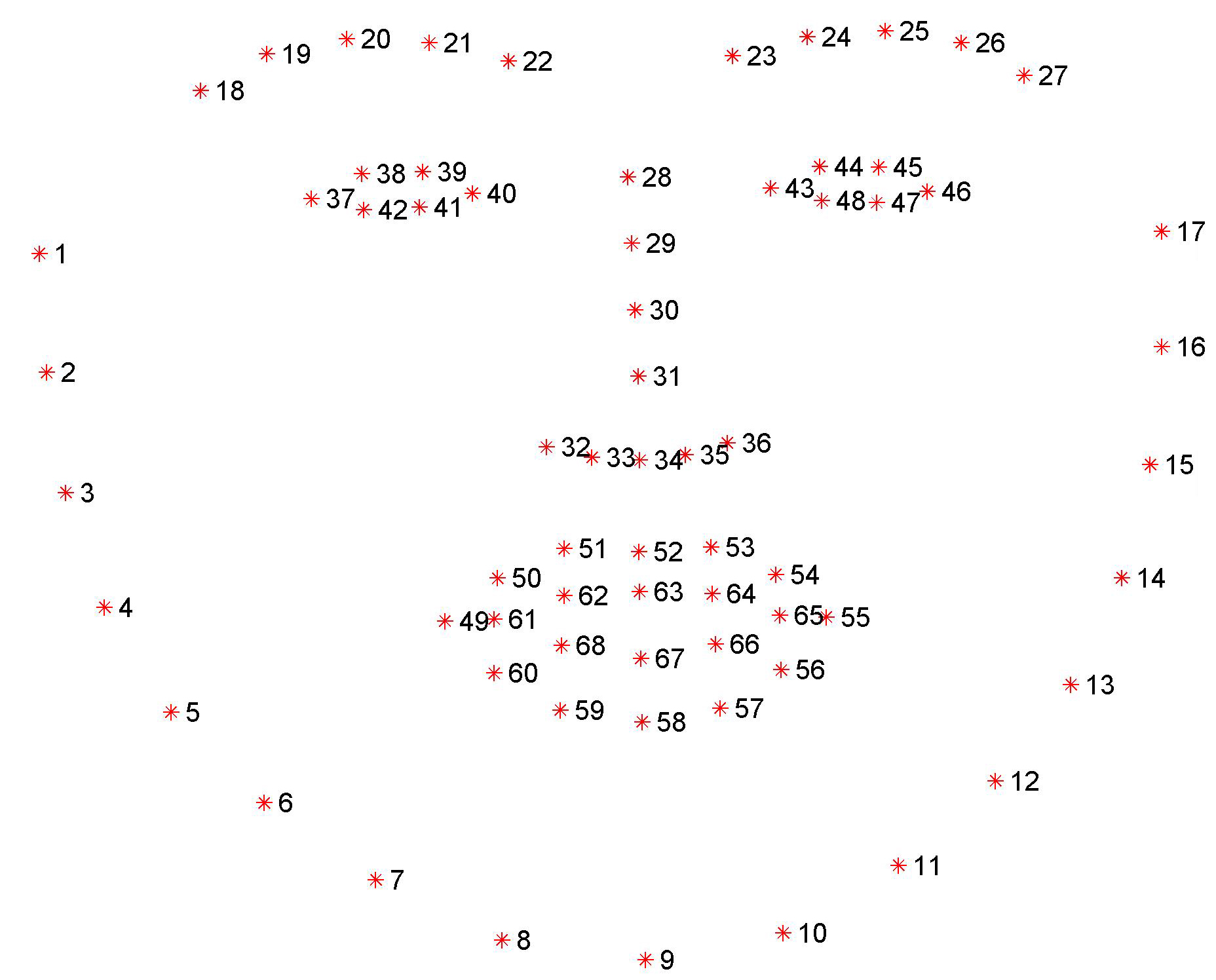}
\caption{Facial Landmarks.}
\end{figure}

\subsubsection{Distance}
Head motion and facial expressions combined can give substantial information regarding the behavior of a person. Their temporal information may convey information regarding effective state of the person and hence one may find correlation among them. This was implemented as features extracted on the facial regions, namely, eyes, mouth, eyebrows and head.\\
The data was sub-sampled as changes between two consecutive frames would be negligible. Hence one in every three frames was used. Initially affine transformation was applied on the facial points over all the frames to get rid of unwanted scale, translation and rotational variance. The reference points used were taken from a particular frame, because the points for that frame when plotted gave a perfectly aligned face with respect to the frame. And this was observed for all the videos.  \\
For left eye, distance between points \{37, 40\} was used as horizontal distance, and between \{38, 42\} for vertical distance. Similarly for right eye, \{43, 46\}, \{44, 48\} for horizontal and vertical distance respectively. For mouth, points \{52, 58\} were used for vertical distance. But for horizontal distance, average of the distance between two pair of points \{55, 49\}, \{65, 61\} was used. For head, horizontal distance was the average of distance between two pair of points-\{2, 16\} and \{4, 14\}, and vertical distance was the average of \{22, 8\} and \{23, 10\}. Since the motion of the two eyebrows occur simultaneously, hence their horizontal and vertical motion was calculated together. For the horizontal motion, average of the distance between two pair of points- \{22, 23\} and \{27, 18\} was used. The vertical motion of the eyebrows was judged with respect to a reference point that hardly moves in the facial expressions, i.e., nose tip. Hence for vertical motion, average of the distance between pairs- \{31, 25\}, \{31, 20\} was used. Hence these distances were calculated for all the frames and were stored in a vector. Hence this resulted in a total of 10 vectors per video.  These vectors were individually normalized by its sum to account for the different type of faces, i.e., a person may have long face or small eyes, but it should not affect our results. In this way normalization removed the unwanted effects that the characteristics of a person's face may have. \\
The resultant vectors produced were further processed using Gaussian Mixture Model (GMM) for producing bag of words. Then the fisher vectors \cite{fish1} \cite{fish2} were extracted for each video from the clusters produced using GMM.\\

GMM model gives the probability of points belonging to a cluster. According to the model the, probability is normally distributed around the mean/centroid of the cluster.
The resultant vectors produced from above approach were further classified into 64 clusters using K-means. Then the clusters formed by K-means were used as inputs for the initialization of the EM (expectation maximization) for forming the GMM.
 
\subsubsection{Emotions, AUs, Gaze and Pose}
Statistical features, namely minimum, maximum, mean, mode, median, range, mean deviation, variance, standard deviation, skewness, and kurtosis were calculated for pre extracted AU, gaze and pose features provided in the given data set. These statistical features were also used for depression analysis.

\subsubsection{Blink Rate}
Blink rate was calculated using the 2D facial landmarks. First, the region of points enclosing the eye (points 37-42, of left eye) were taken. For the entire number of frames given, the area of the polygon made by those points was calculated. This way, the data eye area VS frame data was obtained. The given data was plotted to obtain an idea of blink rate. Now to obtain approximate area of open eye, mode of the total number of areas (per frame) was obtained and considered as area of open eye. For the close eye area, the minimum of the area of random 1000 frames was taken.\\
A blink was considered if the area covered by the eye points is less than 90 percent of the area with eye opened. This way the number of blinks were calculated over the entire number of frames. The blink frequency was calculated by dividing the number of blinks by corresponding duration of the interview.
The sample of the graph of eye area VS frames is shown below.
\captionsetup{singlelinecheck=on}
\begin{figure}[htp]
\centering
\includegraphics[width=8cm]{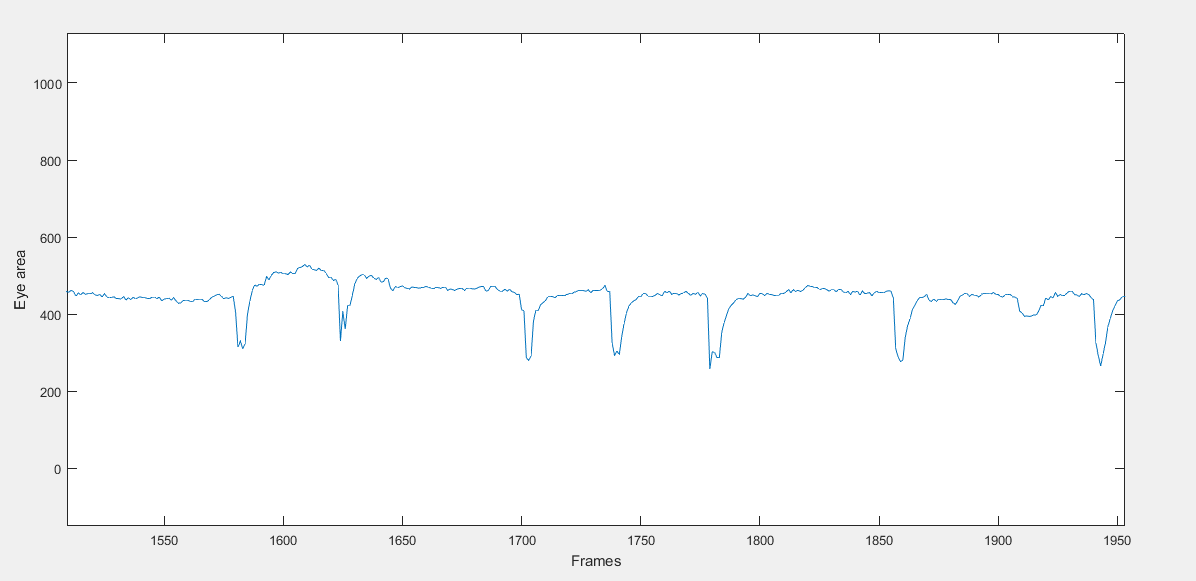}
\caption{eye area VS frames.}
\end{figure}

\subsubsection{Other Features}
 For visual features, Motion History image that converts the motion into grayscale image, with the most recent movement shown by pixels with highest grayscale value and earliest by pixels with least grayscale value \cite{ptucha}, was also computed on the facial 2D landmarks. The in between motion depicted by gradually increasing grayscale value, with 255 for most recent activity. HOG and LBP features were extracted on the resulting MHI formed. But the results obtained from classification on the HOG and LBP features extracted, were vague and inaccurate. Hence this approach is not discussed in detail.

\subsection{Audio Features}
The audio data provided by AVEC consisted of an audio file of entire interview with the participant, pre-extracted features using the COVAREP toolbox at 10-ms intervals over the entire recording (F0, NAQ, QOQ, H1H2, PSP, MDQ, peak Slope, Rd, Rd conf, MCEP 0-24, HMPDM 1-24, HMPDD 1-12, and Formants 1-3), Transcript file containing speaking times and values of participant and the virtual interviewer, and a 
formant file. The given audio file was processed so as to obtain the voice of the participant only. Hence from the audio file, the voice of the participant was isolated using the speaking times giveb in transcript file. \\
Two sets of features were calculated for audio modality. The first set consists of statistical features for low level descriptors as shown in the following Table I. These features were calculated using the inbuilt functions in matlab. These low level descriptors have been pre extracted and provided with the data set in the covarep file.

\begin{table}[h!]
\centering
\begin{tabular}{ |p{4cm}|p{4cm}| } 
\hline

Low Level Descriptors & Statistical features  \\
\hline
normalized F0, NAQ, QOQ, H1H2, PSP, MDQ, peak Slope, Rd, Rd conf, MCEP 0-24, HMPDM 1-24, HMPDD 1-12 & mean, min, skewness, kurtosis, standard deviation, median, peak-magnitude to root-mean-square ratio, root mean square level, interquartile range \\ 
\hline
\end{tabular}
\caption{Statistical descriptors calculated from the pre-extracted audio features.}

\end{table}

The second set of audio features consisted of Discrete Cosine Transform (DCT) coefficients for each descriptor in the first column of Table 1. The first 10 values of the DCT were retained to reduce the complexity of processing data. Finally, the two sets of features were concatenated and used for further analysis.

\subsection{Text Features}
Features were also extracted from the verbal responses of the participant as given in the transcript file. Total number of sentences, words spoken by the participant, average words spoken in each sentence, ratio of the laughter count to the total number of words spoken by participant, ratio of the depression related words to the total number of words spoken by the participant, are the text features extracted. The total number of sentences and words spoken were normalized by duration of the video. Referring to the mentioned paper \cite{ellgring}, slow and less amount of speech, longer speech pauses and brief answers are manifestations of depression. For the negative/depression related words, a dictionary of about 720 words was constructed manually using online resources \cite{maple}, \cite{ghosh}. Another seven features were extracted using the Affective Norms for English Words ratings (ANEW) \cite{bradley}. Mean and Standard Deviation of pleasure, arousal and dominance ratings, word frequency for each word spoken by participant were stored. Eventually mean of these features were taken over all the words, giving total of seven features.

\section{Classification}
SVM and Neural Network were used as classification algorithms. SVM classifier was applied separately on all the above extracted features. For each of the above features, eight models were trained, with each model giving score 0 ,1, 2 or 3 as intensities of PHQ8\_NoInterest, PHQ8\_Depressed, PHQ8\_Sleep, PHQ8\_Tired, PHQ8\_Appetite, PHQ8\_Failure, PHQ8\_Concentrating and PHQ8\_Moving. Results of all the eight models were added to obtain the final predicted PHQ8\_score \cite{PHQ8} out of 24.  \\
Neural Network was applied only on the Fisher vectors calculated because the dimension of other feature vectors for a video were very small and hence application of layers to such small dimension did not make sense. Also it was checked experimentally and the results were not good. Hence for Fisher vectors, similar to SVM, 8 classification networks were trained, whose sum gave the net score out of 24. Another regression neural network model was also trained that gave the combined output of all the 8 PHQ8 labels. The model was constructed with 8 nodes as output of the last layer with each node giving value between 0-3 supposedly. The score of all 8 nodes obtained was first rounded to nearest integer value. Hence the sum of all the 8 nodes gave score out of 24 that was used to judge depression. In this model, interlink was kind of set between different PHQ8 labels and was expected to give better results. But the network with 6 layers gave best RMSE 6.85.
\section{Experimental Results}

\subsection{Support Vector Machine (SVM)}
Initially the default parameters were used for the SVM classifier. But the training accuracy obtained with default parameters was very less. Hence a K-fold cross validation was applied to obtain the optimum values of cost and gamma for the classifier for each of the features extracted above. Cross validation performed was 5 fold, applied using the python script. With cross validation, model was able to fit the training data for a number of values of cost and gamma. Eventually those values of cost and gamma were selected for which accuracy on development set was maximum. The best results along with SVM parameters, obtained for each of the above, i.e , audio, text, head pose and fisher vectors on validation dataset are displayed in Table 2. The results for other modalities are not listed as they were not even close to the baseline. Best RMSE on validation set obtained for Histogram of Oriented Gradients (HOG) on MHI was 7.72. Similar RSME was obtained for Local Binary Pattern (LBP), statistical features of AU, gaze pose and blink rate.
\begin{table}[h!]
\centering
\begin{tabular}{ |p{1cm}|p{1cm}|p{1cm}|p{1cm}|p{1cm}|p{1cm}| } 
\hline

\textbf{features} & \textbf{RMSE} & \textbf{MAE} & \textbf{kernel} & \textbf{cost} & \textbf{gamma}  \\
\hline
Audio & 6.32 & 4.4 & Linear & 9 & -7 \\
\hline
Text & 6.45 & 5.0 & Radical & 9 & -7 \\
\hline
Head & 6.39 & 5.08 & Cyclic & 5 & 1 \\
\hline
fisher & 6.46 & 4.91 & radical & 5 & 1 \\
\hline
\end{tabular}
\caption{Results obtained with SVM on validation set.}

\end{table}

All these results are on the development set.
The results obtained were better than the baseline (Table III) on the development set. For test set, only one submission was made till date, whose result, as mentioned in the Table is not that accurate.

\begin{table}[h!]
\centering
\begin{tabular}{ |p{1.6cm}|p{1.6cm}|p{1.6cm}|p{1.6cm}| } 
\hline

\textbf{Partition} & \textbf{Modality} & \textbf{RMSE} & \textbf{MAE}   \\
\hline
Development & Audio & 6.74 & 5.36 \\
\hline
Development & Video & 7.13 & 5.8 \\
\hline
Development & Audio-Video & 6.62 & 5.52 \\
\hline
Test & Audio & 7.78 & 5.72 \\
\hline
Test & Video & 6.97 & 6.12 \\
\hline
Test & Audio-Video & 7.05 & 5.66 \\
\hline
\end{tabular}
\caption{Provided Baseline Results.}

\end{table}

\subsection{Neural Networks}
For all the eight models, a total of 6 layers neural network was used. Dropout was added between most layers to prevent overfitting. The range of dropout was kept between 0.2-0.5. The number of dropouts and dimensions of layers were manually found so as to keep the RMSE and MAE of each of the network minimum. For all the networks, `Adam' optimizer was used over `SGD' as it was faster as well as efficient. The best RMS and MAE for both optimizers is given in Table IV.

\begin{table}[h!]
\centering
\begin{tabular}{ |p{2cm}|p{2cm}|p{2cm}| } 
\hline

\textbf{} & \textbf{RMSE} & \textbf{MAE} \\
\hline
Adam & 6.342 & 5.085 \\
\hline
SGD & 8.473 & 6.142 \\
\hline
\end{tabular}
\caption{Neural Network classification Result on validation set.}

\end{table}

With the results obtained on the features from both the approaches, SVM and Neural Networks, Decision Level Fusion was applied on the results of different modalities. Outputs of Audio and Text were fused together (Table 4) and Head pose and Fisher together (Table 5). This was so done as the dimensionality of Audio-Text and Fisher-Head pose were close. Since RMSE for each of them were nearly same, so the weights were decided experimentally as mentioned in the Table V and Table VI.

\begin{table}[h!]
\centering
\begin{tabular}{ |p{1.1cm}|p{1.1cm}|p{1.1cm}|p{1.1cm}|p{1.1cm}|p{1.1cm}| } 
\hline

\textbf{Weight(aud)} & \textbf{Weight(text)} & \textbf{RMSE(dev)} & \textbf{MAE(dev)} & \textbf{RMSE(test)} & \textbf{MAE(test)}  \\
\hline
\textbf{0.5} & \textbf{0.5} & \textbf{5.593} & \textbf{4.3714} & \textbf{7.631} & \textbf{6.2766}  \\
\hline
0.6 & 0.4 & 5.789 & 4.314 & & \\
\hline
0.7 & 0.3 & 5.901 & 4.3714 & & \\
\hline
0.8 & 0.2 & 6.009 & 4.285 & & \\
\hline
0.4 & 0.6 & 5.684 & 4.428 & & \\
\hline
0.3 & 0.7 & 5.786 & 4.457 & & \\
\hline
0.2 & 0.8 & 5.916 & 4.600 & & \\
\hline
\end{tabular}
\caption{Fusion of Audio and text features.}

\end{table}

\begin{table}[h!]
\centering
\begin{tabular}{ |p{1.7cm}|p{1.7cm}|p{1.7cm}|p{1.7cm}| } 
\hline

\textbf{Weight(head)} & \textbf{Weight(fisher)} & \textbf{RMSE} & \textbf{MAE} \\
\hline
\textbf{0.5} & \textbf{0.5} & \textbf{5.744} & \textbf{4.3714} \\
\hline
0.6 & 0.4 & 5.789 & 4.314 \\
\hline
0.7 & 0.3 & 5.901 & 4.3714 \\
\hline
0.8 & 0.2 & 6.009 & 4.285 \\
\hline
0.4 & 0.6 & 5.855 & 4.514 \\
\hline
0.3 & 0.7 & 5.87 & 4.514 \\
\hline
0.2 & 0.8 & 6.00 & 4.657 \\
\hline
\end{tabular}
\caption{Fusion of Head pose and fisher features  on validation set.}

\end{table}

All these results are on the development set. On the training set, all models fit quite accurately and gave very less RMSE and MAE for all the cases. Eventually equal weights were given to both the modalities as it gave least RMSE for both the fused models.
Also on fusion of all the four modalities with equal weights given to all, gave the best RMSE and MAE. (Table VII)

\begin{table}[h!]
\centering
\begin{tabular}{ |p{1.1cm}|p{1.1cm}|p{1.1cm}|p{1.1cm}|p{1.1cm}|p{1.1cm}| } 
\hline

\textbf{audio} & \textbf{text} & \textbf{fisher} & \textbf{head} & \textbf{RMSE} & \textbf{MAE}  \\
\hline
\textbf{0.25} & \textbf{0.25} & \textbf{0.25} & \textbf{0.25} & \textbf{5.4143} & \textbf{4.1714}  \\
\hline
\end{tabular}
\caption{Equal weight to results of all four modalities on validation set.}

\end{table}

Another fusion technique applied was to maximize the outputs of both the modalities. With each output compared for both the modalities, maximum among them was taken. This technique was applied across all four modalities also. (Table VIII)

\begin{table}[h!]
\centering
\begin{tabular}{ |p{2cm}|p{2cm}|p{2cm}| } 
\hline

\textbf{Features} & \textbf{RMSE} & \textbf{MAE} \\
\hline
Audio-text & 5.983 & 4.3714 \\
\hline
fisher-head & 5.3799 & 4.6571 \\
\hline
Audio-text-fisher-head & \textbf{5.3586} & \textbf{4.3714} \\
\hline
\end{tabular}
\caption{Fusion Classification results on validation set.}

\end{table}

The results given for test were after training on both train and development sets.

\section{Conclusion}
 Behavior of a depressed person shows relative change in terms of speech pattern, facial expressions and head movement when compared to a non-depressed person. In this paper, we introduced depression recognition task through visual, audio and text features using SVM and neural networks as classifiers. GMM clustering and fisher vectors were calculated on the relative distance of the facial regions. Facial Regions used in recording the relative distance of certain points are the ones mostly involved in facial expressions like smiling, laughing, and other visible emotions. Head pose, statistical descriptors on gaze, pose and blinking rate were also calculated. Verbal responses of the person coded in the form of text (sentences, words, negative words) and audio (low level) features hold the information regarding the behavior of the person. \\
The features extracted were trained on SVM. Results from audio, fisher vectors and text features, both individually and combined outperformed the baseline results on validation data set (Table 7). Fisher vector features were also classified using Neural Networks and also crossed the baseline results on validation data set (Table 7). Further decision fusion in the form of mean and maximizing the outputs was purely experimental. Results obtained on maximizing the outputs of all four modalities trained on SVM and taking mean of the outputs of different modalities were the best. The better accuracy obtained may have removed unwanted variance on the outputs giving desired results.\\
Since the fused outputs improved accuracies, future work would be devoted towards fusing the outputs in more generic manner. Also the models that gave results overfit on the training data, hence parameters search would be to prevent overfitting.

\section{Acknowledgement}
We would like to express our gratitude to our teacher and organizers of the AVEC 2017, who gave us the golden opportunity to participate in this project (AVEC 2017). The project helped us inculcate better understanding of computer vision, machine learning and even basic concepts of deep learning.

\ifCLASSOPTIONcaptionsoff
  \newpage
\fi


\begin{thebibliography}{9}
\balance
\bibitem{depressionarticle} 
Depression article on\\
\url{https://www.nimh.nih.gov/health/topics/depression/index.shtml}
\bibitem{pedersen}
J. Pedersen et al., “An ethological description of depression,”
Acta psychiatrica scandinavica, vol. 78, no. 3, pp. 320–330,
1988.
\bibitem{fossi}
L. Fossi, C. Faravelli, and M. Paoli, “The ethological ap-
proach to the assessment of depressive disorders,” The Jour-
nal of nervous and mental disease, vol. 172, no. 6, pp. 332–
341, 1984.
\bibitem{waxer}
P. Waxer et al., “Nonverbal cues for depression.” Journal of Abnormal Psychology, vol. 83, no. 3, p. 319, 1974.
\bibitem{AVEC16}
Valstar, Michel, et al. "Avec 2016: Depression, mood, and emotion recognition workshop and challenge." Proceedings of the 6th International Workshop on Audio/Visual Emotion Challenge. ACM, 2016. \\
\bibitem{AVEC17}
Ringeval, Fabien, et al. "AVEC 2017–Real-life Depression, and Affect Recognition Workshop and Challenge." (2017). \\
\bibitem{asim}
Jan, Asim, et al. "Automatic depression scale prediction using facial expression dynamics and regression." Proceedings of the 4th International Workshop on Audio/Visual Emotion Challenge. ACM, 2014.
\bibitem{acoustic}
Low, L. "Detection of clinical depression in adolescents' using acoustic speech analysis." (2011).
\bibitem{sturim}
Sturim, Douglas, et al. "Automatic detection of depression in speech using gaussian mixture modeling with factor analysis." Twelfth Annual Conference of the International Speech Communication Association. 2011.
\bibitem{sharifa}
Alghowinem, Sharifa, et al. "Head pose and movement analysis as an indicator of depression." Affective Computing and Intelligent Interaction (ACII), 2013 Humaine Association Conference on. IEEE, 2013.
\bibitem{tzirakis}
Tzirakis, Panagiotis, et al. "End-to-End Multimodal Emotion Recognition using Deep Neural Networks." arXiv preprint arXiv:1704.08619 (2017).
\bibitem{pampou}
Pampouchidou, Anastasia, et al. "Depression Assessment by Fusing High and Low Level Features from Audio, Video, and Text." Proceedings of the 6th International Workshop on Audio/Visual Emotion Challenge. ACM, 2016.
\bibitem{ptucha}
Ptucha, Raymond, and Andreas Savakis. "Towards the Usage of Optical Flow Temporal Features for Facial Expression Classification." Advances in Visual Computing (2012): 388-397.
\bibitem{ellgring}
Ellgring, Heiner. Non-verbal communication in depression. Cambridge University Press, 2007.
\bibitem{maple}
mr. L. (Maple Canada) Depression Vocabulary, Depression Word List.
\bibitem{ghosh}
Ghosh, Sayan, Moitreya Chatterjee, and Louis-Philippe Morency. "A multimodal context-based approach for distress assessment." Proceedings of the 16th International Conference on Multimodal Interaction. ACM, 2014.
\bibitem{bradley}
Bradley, Margaret M., and Peter J. Lang. Affective norms for English words (ANEW): Instruction manual and affective ratings. Technical report C-1, the center for research in psychophysiology, University of Florida, 1999.
\bibitem{symp}
Katon, Wayne, and Mark D. Sullivan. "Depression and chronic medical illness." J Clin Psychiatry 51.Suppl 6 (1990): 3-11.
\bibitem{Cumm}
Cummins, Nicholas, et al. "Diagnosis of depression by behavioural signals: a multimodal approach." Proceedings of the 3rd ACM international workshop on Audio/visual emotion challenge. ACM, 2013.
\bibitem{fish1}
Oneata, Dan, Jakob Verbeek, and Cordelia Schmid. "Action and event recognition with fisher vectors on a compact feature set." Proceedings of the IEEE international conference on computer vision. 2013.
\bibitem{fish2}
Dhall, Abhinav, and Roland Goecke. "A temporally piece-wise fisher vector approach for depression analysis." Affective Computing and Intelligent Interaction (ACII), 2015 International Conference on. IEEE, 2015.
\bibitem{PHQ8}
Kroenke, Kurt, et al. "The PHQ-8 as a measure of current depression in the general population." Journal of affective disorders 114.1 (2009): 163-173.
\bibitem{Le Yang}
Yang, Le, et al. "Decision Tree Based Depression Classification from Audio Video and Language Information." Proceedings of the 6th International Workshop on Audio/Visual Emotion Challenge. ACM, 2016.
\end{thebibliography}
\end{document}